\def\year{2019}\relax
\documentclass[letterpaper]{article} 
\usepackage{aaai19}  
\usepackage{times}  
\usepackage{helvet}  
\usepackage{courier}  
\usepackage{url}  
\usepackage{graphicx}  
\usepackage{comment}
\usepackage{amssymb}
\usepackage{amsmath}
\usepackage{amsfonts}
\usepackage{amsthm,bbm}
\usepackage{color}
\usepackage[para,online,flushleft]{threeparttable}
\newtheorem{theorem}{Theorem}
\newtheorem{assumption}{Assumption}
\newtheorem{corollary}{Corollary}
\newtheorem{example}{Example}

\newtheorem{lemma}{Lemma}
\newtheorem{definition}{Definition}
\newcommand{\argmax}{\mathop{\rm argmax}\limits}
\newcommand{\argmin}{\mathop{\rm argmin}\limits}
\newcommand{\indep}{\mathop{\perp\!\!\!\!\perp}}
\newcommand{\citet}[1]
{\citeauthor{#1} \shortcite{#1}}
\newcommand{\citep}{\cite}

\allowdisplaybreaks
\title{Efficient Counterfactual Learning from Bandit Feedback}
\author{Yusuke Narita\\ Yale University \\yusuke.narita@yale.edu 
\And Shota Yasui\\ CyberAgent Inc. \\yasui\_shota@cyberagent.co.jp 
\And Kohei Yata\\ Yale University \\kohei.yata@yale.edu
}
\begin{document}
%
\maketitle
\begin{abstract}
\noindent What is the most statistically efficient way to do off-policy optimization with batch data from bandit feedback? 
For log data generated by contextual bandit algorithms, we consider offline estimators for the expected reward from a counterfactual policy. 
Our estimators are shown to have lowest variance in a wide class of estimators, achieving variance reduction relative to standard estimators. 
We then apply our estimators to improve advertisement design by a major advertisement company. 
Consistent with the theoretical result, our estimators allow us to improve on the existing bandit algorithm with more statistical confidence compared to a state-of-the-art benchmark.
\end{abstract}
\section{Introduction}

Interactive bandit systems (e.g. personalized education and medicine, ad/news/recommendation/search platforms) produce log data valuable for evaluating and redesigning the systems. 
For example, the logs of a news recommendation system record which news article was presented and whether the user read it, giving the system designer a chance to make its recommendation more relevant. 
Exploiting log data is, however, more difficult than conventional supervised machine learning: 
the result of each log is only observed for the action chosen by the system (e.g. the presented news) but not for all the other actions the system could have taken. 
Moreover, the log entries are biased in that the logs over-represent actions favored by the system. 

A potential solution to this problem is an A/B test that compares the performance of counterfactual systems. 
However, A/B testing counterfactual systems is often technically or managerially infeasible, since deploying a new policy is time- and money-consuming, and entails a risk of failure.

This leads us to the problem of \textit{counterfactual (off-policy) evaluation and learning}, where one aims to use batch data collected by a logging policy to estimate the value of a counterfactual policy or algorithm without employing it \citep{li2010contextual,Strehl2010,li2011unbiased,Li2012,bottou2013counterfactual,Swaminathan2015,Swaminathan2015b,wang2016optimal,swaminathan2017off}. 
Such evaluation allows us to compare the performance of counterfactual policies to decide which policy should be deployed in the field. 
This alternative approach thus solves the above problem with the naive A/B test approach. \\

\textbf{Method.} 
For off-policy evaluation with log data of bandit feedback, this paper develops and empirically implements a variance minimization technique. 
Variance reduction and statistical efficiency are important for minimizing the uncertainty we face in decision making.  
Indeed, an important open question raised by \citet{li2015offline} is how to achieve ``statistically more efficient (even optimal) offline estimation" from batch bandit data. 
This question motivates a set of studies that bound and characterize the variances of particular estimators \citep{Dudik2014,li2015toward,thomas2015high,munos2016safe,thomas2016data,agarwal2017effective}. 

We study this statistical efficiency question in the context of offline policy value estimation with log data from a class of contextual bandit algorithms. 
This class includes most of the widely-used algorithms such as contextual $\epsilon$-Greedy and Thompson Sampling, as well as their non-contextual analogs and random A/B testing. 
We allow the logging policy to be unknown, degenerate (non-stochastic), and time-varying, all of which are salient in real-world bandit applications. 
We also allow the evaluation target policy to be degenerate, again a key feature of real-life situations. 

We consider offline estimators for the expected reward from a counterfactual policy. 
Our estimators can also be used for estimating the average treatment effect. 
Our estimators are variations of well-known inverse probability weighting estimators (\citet{horvitz1952generalization}, \citet{rosenbaum1983central}, and modern studies cited above) except that we use an \textit{estimated} propensity score (logging policy) even if we know the true propensity score. 
We show the following result, building upon \citet{Bickel1993}, \citet{HIR2003}, and \citet{Ackerberg2014} among others:
\begin{quote}
\textbf{Theoretical Result 1.} Our estimators minimize the variance among all reasonable estimators. 
More precisely, our estimators minimize the asymptotic variance among all ``asymptotically normal" estimators (in the standard statistical sense defined in Section \ref{estimation}). 
\end{quote}
We also provide estimators for the asymptotic variances of our estimators, thus allowing analysts to calculate the variance in practice. 
In contrast to Result 1, we also find: 
\begin{quote}
\textbf{Theoretical Result 2.} Standard estimators using the true propensity score (logging policy) have larger asymptotic variances than our estimators. 
\end{quote}

\noindent Perhaps counterintuitively, therefore, the policy-maker should use an estimated propensity score even when she knows the true one. \\

\textbf{Application.} We empirically apply our estimators to evaluate and optimize the design of online advertisement formats. 
Our application is based on proprietary data provided by CyberAgent Inc., the second largest Japanese advertisement company with about 6 billion USD market capitalization (as of November 2018). 
This company uses a contextual bandit algorithm to determine the visual design of advertisements assigned to users. 
Their algorithm produces logged bandit data. 

We use this data and our estimators to optimize the advertisement design for maximizing the click through rates (CTR). 
In particular, we estimate how much the CTR would be improved by a counterfactual policy of choosing the best action (advertisement) for each context (user characteristics). 
We first obtain the following result: 

\begin{quote}
\textbf{Empirical Result A.} Consistent with Theoretical Results 1-2, our estimators produce narrower confidence intervals about the counterfactual policy's CTR than a benchmark using the known propensity score \cite{Swaminathan2015b}. 
\end{quote}

\begin{figure}
\begin{center}
 \includegraphics[width=85mm]{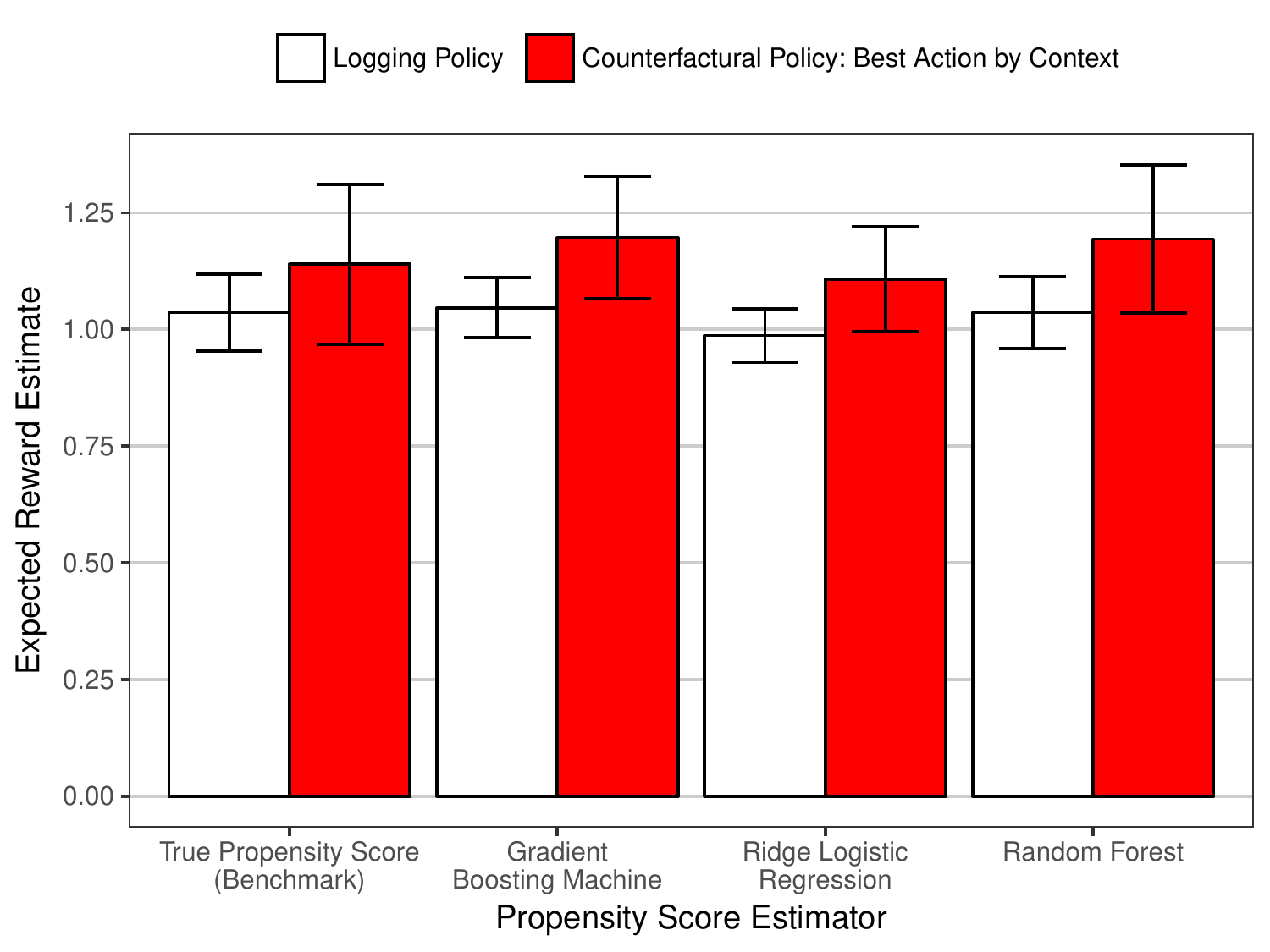}
\end{center}
\caption{Improving Ad Design with Lower Uncertainty}\label{figure}\par
~\par
         \fontsize{9.0pt}{10.0pt}\selectfont \textit{Notes}: This figure shows estimates of the expected CTRs of the logging policy and a counterfactual policy of choosing the best action for each context. 
         CTRs are multiplied by a constant for confidentiality reasons. 
         We obtain these estimates by the ``self-normalized inverse probability weighting estimator" using the true propensity score (benchmark thanks to \citet{Swaminathan2015b}) or estimated propensity scores (our proposal), both of which are defined and analyzed in Section \ref{estimation}. 
         Bars indicate 95\% confidence intervals based on our asymptotic variance estimators developed in Section \ref{section:variance}.
\end{figure}

\noindent This result is reported in Figure \ref{figure}, where the confidence intervals using ``True Propensity Score (Benchmark)" are wider than other confidence intervals using propensity scores estimated either by the Gradient Boosting, Random Forest, or Ridge Logistic Regression. 

Thanks to this variance reduction, we conclude that the logging policy's CTR is below the confidence interval of the hypothetical policy of choosing the best advertisement for each context. 
This leads us to obtain the following bottomline: 

\begin{quote}
\textbf{Empirical Result B.} Unlike the benchmark estimator, our estimator predicts the hypothetical policy to statistically significantly improve the CTR by 10-15\% (compared to the logging policy). 
\end{quote}

\noindent Empirical Results A and B therefore show that our estimator can substantially reduce uncertainty we face in real-world policy-making. 

\section{Setup}\label{model}

\subsection{Data Generating Process}\label{dgp}

We consider a general multi-armed contextual bandit setting. 
There is a set of $m+1$ \textit{actions} (equivalently, \textit{arms} or \textit{treatments}), ${\cal A}=\{0,...,m\}$, that the decision maker can choose from.
Let $Y(\cdot): {\cal A}\rightarrow \mathbb{R}$ denote a potential reward function that maps actions into rewards or outcomes, where $Y(a)$ is the reward when action $a$ is chosen (e.g., whether an advertisement as an action results in a click).
Let $X$ denote \textit{context} or \textit{covariates} (e.g., the user's demographic profile and browsing history) that the decision maker observes when picking an action. 
We denote the set of contexts by ${\cal X}$.
We think of $(Y(\cdot),X)$ as a random vector with unknown distribution $G$.

We consider log data coming from the following data generating process (DGP), which is similar to those used in the literature on the offline evaluation of contextual bandit algorithms \citep{li2010contextual,Strehl2010,li2011unbiased,Li2012,Swaminathan2015,Swaminathan2015b,swaminathan2017off}.
We observe data $\{(Y_t,X_t,D_t)\}_{t=1}^T$ with $T$ observations. 
$D_t\equiv(D_{t0},...,D_{tm})'$ where $D_{ta}$ is a binary variable indicating whether action $a$ is chosen in round $t$. 
$Y_t$ denotes the reward observed in round $t$, i.e., $Y_t\equiv \sum_{a=0}^{m}D_{ta}Y_t(a)$. 
$X_t$ denotes the context observed in round $t$.

A key feature of our DGP is that the data $\{(Y_t,X_t,D_t)\}_{t=1}^T$ are divided into $B$ batches, where different batches may use different choice probabilities (propensity scores). 
Let $X_t^b\in\{1,2,...,B\}$ denote a random variable indicating the batch to which round $t$ belongs.
We treat this batch number as one of context variables and write $X_t=(\tilde{X}_t,X_t^b)$, where $\tilde X_t$ is the vector of context variables other than $X_t^b$. 

Let $p_t=(p_{t0},...,p_{tm})'\in \Delta({\cal A})$ denote the potentially unknown probability vector indicating the probability that each action is chosen in round $t$. 
Here $\Delta({\cal A})\equiv\{(p_a)\in\mathbb{R}^{m+1}_{+}| \sum_a p_a=1\}$ with $p_a$ being the probability that action $a$ is chosen. 
A {\it contextual bandit algorithm} is a sequence $\{F_b\}_{b=1}^B$ of distribution functions of choice probabilities $p_t$ conditional on $\tilde X_t$, where $F_b: \tilde{\cal X}\rightarrow \Delta(\Delta({\cal A}))$ for $b\in\{1,2,...,B\}$ and $\tilde{\cal X}$ is the support of $\tilde X$, where $\Delta(\Delta({\cal A}))$ is the set of distributions over $\Delta({\cal A})$.
$F_b$ takes context $\tilde X_t$ as input and returns a distribution of probability vector $p_t$ in rounds of batch $b$.
$F_b$ can vary across batches but does not change across rounds within batch $b$. 
We assume that the log data are generated by a contextual bandit algorithm $\{F_b\}_{b=1}^B$ as follows:

\begin{itemize}
\item In each round $t=1,...,T$, $(Y_t(\cdot),X_t)$ is i.i.d. drawn from distribution $G$. 
Re-order round numbers so that they are monotonically increasing in their batch numbers $X_t^b$. 
\item In each round $t$ within batch $b\in\{1,2,...,B\}$ and given $\tilde X_t$, probability vector $p_t=(p_{t0},...,p_{tm})'$ is drawn from $F_b(\cdot|\tilde X_t)$. 
Action is randomly chosen based on probability vector $p_t$, creating the action choice $D_{t}$ and the associated reward $Y_t$. 
\end{itemize}

\noindent Here, the contextual bandit algorithm $\{F_b\}_{b=1}^B$ and the realized probability vectors $\{p_t\}_{t=1}^T$ may or may not be known to the analyst. 
We also allow for the realization of $p_t$ to be degenerate, i.e., a certain action may be chosen with probability $1$ at a point in time.\\


\textbf{Examples.} This DGP allows for many popular bandit algorithms, as the following examples illustrate. 
In each of the examples below, the contextual bandit algorithm $F_b$ is degenerate and produces a particular probability vector $p_t$ for sure. 

\begin{example}[Random A/B testing]\label{ex:A/B}
	We always choose each action uniformly at random: $p_{ta}=\frac{1}{m+1}$ always holds for any $a\in{\cal A}$ and any $t=1,...,T$.
\end{example}

In the remaining examples, at every batch $b$, the algorithm uses the history of observations from the previous $b-1$ batches to estimate the mean and the variance of the potential reward under each action conditional on each context: $\mu(a|x)\equiv\mathbb E[Y(a)|\tilde X=x]$ and $\sigma^2(a|x)\equiv\mathbb V[Y(a)|\tilde X=x]$.
We denote the estimates using the history up to batch $b-1$ by $\hat \mu_{b-1}(a|x)$ and $\hat \sigma^2_{b-1}(a|x)$. 
See \citet{Li2012} and \citet{Dimakopoulou2017} for possible estimators based on generalized linear models and generalized random forest, respectively.
The initial estimates, $\hat \mu_{0}(a|x)$ and $\hat \sigma^2_{0}(a|x)$, are set to any values. 

\begin{example}[$\epsilon$-Greedy]\label{ex:e-greedy}
	In each round within batch $b$, we choose the best action based on $\hat \mu_{b-1}(a|\tilde X_t)$ with probability $1-\epsilon_b$ and choose the other actions uniformly at random with probability $\epsilon_b$:
	$$
	p_{ta}= \begin{cases}
	1-\epsilon_b & \ \ \ \text{if $a=\argmax_{a'\in {\cal A}}\hat \mu_{b-1}(a'|\tilde X_t)$}\\
	\dfrac{\epsilon_b}{m} & \ \ \ \text{otherwise}.
	\end{cases}
	$$
\end{example}


\begin{example}[Thompson Sampling using Gaussian priors]\label{ex:Thompson}
	In each round within batch $b$, we sample the potential reward $y_t(a)$ from distribution ${\cal N}(\hat \mu_{b-1}(a|\tilde X_t),\hat \sigma^2_{b-1}(a|\tilde X_t))$ for each action, and choose the action with the highest sampled potential reward, $\argmax_{a'\in {\cal A}}y_t(a')$.
	As a result, this algorithm chooses actions with the following probabilities:
	$$
	p_{ta} = \Pr\{a=\argmax_{a'\in {\cal A}}y_t(a')\},
	$$
	where $(y_t(0),...,y_t(m))'\sim {\cal N}(\hat \mu_{b-1}(\tilde X_t),\hat \Sigma_{b-1}(\tilde X_t))$, $\hat \mu_{b-1}(x)=(\hat \mu_{b-1}(0|x),...,\hat \mu_{b-1}(m|x))'$, and
	$$
	\hat \Sigma_{b-1}(x)=
	\begin{pmatrix}
	\hat \sigma^2_{b-1}(0|x) & 0 & 0\\
	0 & \ddots & 0 \\
	0 & 0 & \hat \sigma^2_{b-1}(m|x) \\
	\end{pmatrix}.
	$$
\end{example}

In Examples \ref{ex:e-greedy} and \ref{ex:Thompson}, $p_t$ depends on the random realization of the estimates $\hat \mu_{b-1}(a|x)$ and $\hat \sigma^2_{b-1}(a|x)$, and so does the associated $F_b$.
If the data are sufficiently large, the uncertainty in the estimates vanishes: $\hat \mu_{b-1}(a|x)$ and $\hat \sigma^2_{b-1}(a|x)$ converge to $\mu_{b-1}(a|x)\equiv\mathbb E[Y(a)|\tilde X=x,X^b\le b-1]$ and $\sigma^2_{b-1}(a|x)\equiv\mathbb V[Y(a)|\tilde X=x,X^b\le b-1]$, respectively. 
In this case, $F_b$ becomes nonrandom since it depends on the fixed realizations $\mu_{b-1}(a|x)$ and $\sigma^2_{b-1}(a|x)$.
In the following analysis, we consider this large-sample scenario and assume that $F_b$ is nonrandom.

To make the notation simpler, we put $\{F_b\}_{b=1}^B$ together into a single distribution $F: {\cal X}\rightarrow \Delta(\Delta({\cal A}))$ obtained by $F(\cdot|\tilde X, X^b=b)=F_b(\cdot|\tilde X)$ for each $b\in\{1,2,...,B\}$.
We use this to rewrite our DGP as follows:

\begin{itemize}
	\item In each round $t=1,...,T$, $(Y_t(\cdot),X_t)$ is i.i.d. drawn from distribution $G$.
	Given $X_t$, probability vector $p_t=(p_{t0},...,p_{tm})'$ is drawn from $F(\cdot|X_t)$. 
	Action is randomly chosen based on probability vector $p_t$, creating the action choice $D_{t}$ and the associated reward $Y_t$. 
\end{itemize}

Define
\begin{align*}
p_{0a}(x)\equiv\Pr_{D\sim p, ~p\sim F}(D_{a}=1|X=x)
\end{align*}
for each $a$, and let $p_0(x)=(p_{00}(x),...,p_{0m}(x))'$.
This is the choice probability vector conditional on each context.
We call $p_0(\cdot)$ the {\it logging policy} or the {\it propensity score}.

$F$ is common for all rounds regardless of the batch to which they belong. 
Thus $p_t$ and $D_t$ are i.i.d. across rounds.
Because $(Y_t(\cdot),X_t)$ is i.i.d. and $Y_t=\sum_{a=0}^m D_{ta}Y_t(a)$, each observation $(Y_t,X_t,D_t)$ is i.i.d..
Note also that $D_t$ is independent of $Y_t(\cdot)$ conditional on $X_t$.
We use $p=(p_0,...,p_m)'$ to denote a random vector that has the same distribution as $p_{t}$.

\subsection{Parameters of Interest}


We are interested in using the log data to estimate the expected reward from any given {\it counterfactual policy} $\pi:{\cal X}\rightarrow \bar{\Delta}({\cal A})$, which chooses a distribution of actions given each context:
\begin{align}\label{eq:V^pi}
V^{\pi}&\equiv \mathbb{E}_{(Y(\cdot),X)\sim G}[\sum_{a=0}^mY(a)\pi(a|X)]\nonumber\\
&= \mathbb{E}_{(Y(\cdot),X)\sim G, ~D\sim p_0(X)}[\sum_{a=0}^m Y(a)D_{a}\frac{\pi(a|X)}{p_{0a}(X)}],
\end{align}
where the last equality uses the independence of $D$ and $Y(\cdot)$ conditional on $X$ and the definition of $p_0(\cdot)$.
Here, $\bar{\Delta}({\cal A})\equiv\{(p_a)\in\mathbb{R}^{m+1}| \sum_a p_a\leq 1\}$. 
We allow the counterfactual policy $\pi$ to be degenerate, i.e., $\pi$ may choose a particular action with probability 1. 

Depending on the choice of $\pi$, $V^\pi$ represents a variety of parameters of interest.
When we set $\pi(a|x)=1$ for a particular action $a$ and $\pi(a'|x)=0$ for all $a'\in {\cal A}\backslash\{a\}$ for all $x\in{\cal X}$, $V^\pi$ equals $\mathbb{E}_{(Y(\cdot),X)\sim G}[Y(a)]$, the expected reward from action $a$.
When we set $\pi(a|x)=1$, $\pi(0|x)=-1$ and $\pi(a'|x)=0$ for all $a'\in {\cal A}\backslash\{0,a\}$ for all $x\in{\cal X}$, $V^\pi$ equals $\mathbb{E}_{(Y(\cdot),X)\sim G}[Y(a)-Y(0)]$, the average treatment effect of action $a$ over action $0$.
Such treatment effects are of scientific and policy interests in medical and social sciences.  
Business and managerial interests also motivate treatment effect estimation. 
For example, when a company implements a bandit algorithm using a particular reward measure like an immediate purchase, the company is often interested in treatment effects on other outcomes like long-term user retention.

\section{Efficient Value Estimation}\label{estimation}

We consider the efficient estimation of the expected reward from a counterfactual policy, $V^{\pi}$. 
We consider an estimator consisting of two steps.
In the first step, we nonparametrically estimate the propensity score vector $p_0(\cdot)$ by a consistent estimator.
Possible estimators include machine learning algorithms such as gradient boosting, as well as nonparametric sieve estimators and kernel regression estimators, as detailed in Section \ref{section:pscore}.
In the second step, we plug the estimated propensity $\hat p(\cdot)$ into the sample analogue of expression (\ref{eq:V^pi}) to estimate $V^{\pi}$ (in practice, some trimming or thresholding may be desirable for numerical stability):
$$
\hat V^{\pi}=\frac{1}{T}\sum_{t=1}^T\sum_{a=0}^m Y_tD_{ta}\frac{\pi(a|X_t)}{\hat p_{a}(X_t)}.
$$ 
Alternatively, one can use a ``self-normalized" estimator inspired by \citet{Swaminathan2015b} when $\sum_{a=0}^m\pi(a|x)=1$ for all $x\in {\cal X}$:
$$
\hat V^{\pi}_{SN}=\dfrac{\frac{1}{T}\sum_{t=1}^T\sum_{a=0}^m Y_tD_{ta}\frac{\pi(a|X_t)}{\hat p_{a}(X_t)}}{\frac{1}{T}\sum_{t=1}^T\sum_{a=0}^m D_{ta}\frac{\pi(a|X_t)}{\hat p_{a}(X_t)}}.
$$
\citet{Swaminathan2015b} suggest that $\hat V^{\pi}_{SN}$ tends to be less biased than $\hat V^{\pi}$ in small sample. 
Unlike \citet{Swaminathan2015b}, however, we use the estimated propensity score rather than the true one.

The above estimators estimate a scalar parameter $V^{\pi}$ defined as a function of the distribution of $(Y(\cdot),X)$, on which we impose no parametric assumption. 
Our estimators therefore attempt to solve a semiparametric estimation problem, i.e., a partly-parametric and partly-nonparametric estimation problem.
For this semiparametric estimation problem, we first derive the \textit{semiparametric efficiency bound} on how efficient and precise the estimation of the parameter can be, which is a semiparametric analog of the Cramer-Rao bound \citep{Bickel1993}.
The asymptotic variance of any asymptotically normal estimator is no smaller than the semiparametric efficiency bound. 
Following the standard statistics terminology, we say that estimator $\hat \theta$ for parameter $\theta$ is \textit{asymptotically normal} if $\sqrt{T}(\hat \theta-\theta)\rightsquigarrow {\cal N}(0,\Sigma)$ as $T\rightarrow \infty$, where $\rightsquigarrow$ denotes convergence in distribution, and ${\cal N}(0,\Sigma)$ denotes a normally distributed random variable with mean $0$ and variance $\Sigma$.
We call $\Sigma$ the \textit{asymptotic variance} of $\hat \theta$.
The semiparametric efficiency bound for $\theta$ is a lower bound on the asymptotic variance of asymptotically normal estimators; Appendix \ref{SEB} provides a formal definition of the semiparametric efficiency bound.

We show the above estimators achieve the semiparametric efficiency bound, i.e., they minimize the asymptotic variance among all asymptotically normal estimators. Our analysis uses a couple of regularity conditions. 
We first assume that the logging policy $p_{0}(\cdot)$ ex ante chooses every action with a positive probability for every context.

\begin{assumption}\label{nondegeneracy}
	There exists some $\underline{p}$ such that $0<\underline{p}\le \Pr_{D\sim p, ~p\sim F}(D_{a}=1|X=x)\equiv p_{0a}(x)$ for any $x\in{\cal X}$ and for $a=0,...,m$.
\end{assumption}

\noindent Note that Assumption \ref{nondegeneracy} is consistent with the possibility that the realization of $p_{ta}$ takes on value $0$ or $1$ (as long as it takes on positive values with a positive probability).

We also assume the existence of finite second moments of potential rewards. 

\begin{assumption}\label{finite_variance}
	$\mathbb{E}[Y(a)^2]<\infty$ for $a=0,...,m$.
\end{assumption}

The following proposition provides the semiparametric efficiency bound for $V^{\pi}$. 
All the proofs are in Appendix \ref{proofs}. 

\begin{lemma}[Semiparametric Efficiency Bound]\label{prop:bound:stationary}
	Under Assumptions \ref{nondegeneracy} and \ref{finite_variance}, the semiparametric efficiency bound for $V^{\pi}$, the expected reward from counterfactual policy $\pi$, is
	$$
	\mathbb{E}[\sum_{a=0}^m \mathbb{V}[Y(a)|X]\frac{\pi(a|X)^2}{p_{0a}(X)}+(\theta(X)-V^{\pi})^2],
	$$
	where $\theta(X)=\sum_{a=0}^m \mathbb{E}[Y(a)|X]\pi(a|X)$ is the expected reward from policy $\pi$ conditional on $X$.
\end{lemma}

Lemma \ref{prop:bound:stationary} implies the semiparametric efficiency bounds for the expected reward from each action and for the average treatment effect, since they are special cases of $V^\pi$.

\begin{corollary}\label{corollary:bound:action}
	Suppose that Assumptions \ref{nondegeneracy} and \ref{finite_variance} hold.
	Then, the semiparametric efficiency bound for the expected reward from each action, $\mathbb{E}[Y(a)]$, is
	$$
	\mathbb{E}\bigl[\frac{\mathbb{V}[Y(a)|X]}{p_{0a}(X)}+(\mathbb{E}[Y(a)|X]-\mathbb{E}[Y(a)])^2\bigr].
	$$
	The semiparametric efficiency bound for the average treatment effect, $\mathbb{E}[Y(a)-Y(0)]$, is
	\begin{align*}
	&\mathbb{E}\bigl[\frac{\mathbb{V}[Y(0)|X]}{p_{00}(X)}+\frac{\mathbb{V}[Y(a)|X]}{p_{0a}(X)}\\
	&~~~~~+(\mathbb{E}[Y(a)-Y(0)|X]-\mathbb{E}[Y(a)-Y(0)])^2\bigr].
	\end{align*}
\end{corollary}

Our proposed estimators are two-step generalized-method-of-moment estimators and are asymptotically normal under some regularity conditions, one of which requires that the convergence rate of $\hat p(\cdot)$ be faster than $n^{1/4}$ \cite{Newey1994,chen2007large}.
Given the asympotic normality of the estimators, we find that they achieve the semiparametric efficiency bound, building upon \citet{Ackerberg2014} among others.

\begin{theorem}[Efficient Estimators]\label{prop:estimator:stationary}
	Suppose that Assumptions \ref{nondegeneracy} and \ref{finite_variance} hold and that $\hat p(\cdot)$ is a consistent estimator for $p_0(\cdot)$.
	Then, the variance of $\hat V^{\pi}$ and $\hat V^{\pi}_{SN}$ achieves the semiparametric efficiency bound for $V^{\pi}$ (provided in Lemma \ref{prop:bound:stationary}).
\end{theorem}


\subsection{Inefficient Value Estimation}

In some environments, we know the true $p_0(\cdot)$ or observe the realization of the probability vectors $\{p_t\}_{t=1}^T$.
In this case, an alternative way to estimate $V^\pi$ is to use the sample analogue of the expression (\ref{eq:V^pi}) without estimating the propensity score. 
If we know $p_0(\cdot)$, a possible estimator is
$$
\tilde V^{\pi}=\frac{1}{T}\sum_{t=1}^T\sum_{a=0}^m Y_tD_{ta}\frac{\pi(a|X_t)}{p_{0a}(X_t)}.
$$
If we observe the realization of $\{p_t\}_{t=1}^T$, we may use
$$
\ddot V^{\pi}=\frac{1}{T}\sum_{t=1}^T\sum_{a=0}^m Y_tD_{ta}\frac{\pi(a|X_t)}{p_{ta}}.
$$
When $\sum_{a=0}^m\pi(a|x)=1$ for all $x\in {\cal X}$, it is again possible to use their self-normalized versions: 
$$
\tilde V^{\pi}_{SN}=\dfrac{\frac{1}{T}\sum_{t=1}^T\sum_{a=0}^m Y_tD_{ta}\frac{\pi(a|X_t)}{p_{0a}(X_t)}}{\frac{1}{T}\sum_{t=1}^T\sum_{a=0}^m D_{ta}\frac{\pi(a|X_t)}{p_{0a}(X_t)}}.
$$
$$
\ddot V^{\pi}_{SN}=\dfrac{\frac{1}{T}\sum_{t=1}^T\sum_{a=0}^m Y_tD_{ta}\frac{\pi(a|X_t)}{p_{ta}}}{\frac{1}{T}\sum_{t=1}^T\sum_{a=0}^m D_{ta}\frac{\pi(a|X_t)}{p_{ta}}}.
$$

These intuitive estimators turn out to be less efficient than the estimators with the estimated propensity score, as the following result shows.

\begin{theorem}[Inefficient Estimators]\label{prop:true_pscore:stationary}
	Suppose that the propensity score $p_0(\cdot)$ is known and we observe the realization of $\{p_t\}_{t=1}^T$.
	Suppose also that Assumptions \ref{nondegeneracy} and \ref{finite_variance} hold and that $\hat p(\cdot)$ is a consistent estimator for $p_0(\cdot)$.
	Then, the asymptotic variances of $\tilde V^{\pi}, \ddot V^{\pi}, \tilde V^{\pi}_{SN},$ and $\ddot V^{\pi}_{SN}$ are no smaller than that of $\hat V^{\pi}$ and $\hat V^{\pi}_{SN}$. 
	Generically, $\tilde V^{\pi}, \ddot V^{\pi}, \tilde V^{\pi}_{SN},$ and $\ddot V^{\pi}_{SN}$ are strictly less efficient than $\hat V^{\pi}$ and $\hat V^{\pi}_{SN}$ in the following sense.
	\begin{enumerate}
	\item If $\Pr(\mathbb{E}[Y(a)|X]\frac{\pi(a|X)}{p_{0a}(X)}\neq \theta(X)~\text{for some}~ a)>0$, then the asymptotic variances of $\tilde V^{\pi}$, $\ddot V^{\pi}$, $\tilde V^{\pi}_{SN}$ and $\ddot V^{\pi}_{SN}$ are strictly larger than that of $\hat V^{\pi}$ and $\hat V^{\pi}_{SN}$.
	\item If $\Pr(\mathbb{E}[Y(a)^2|X]\pi(a|X)^2(\mathbb{E}[\frac{1}{p_a}|X]-\frac{1}{p_{0a}(X)})\neq 0~\text{for some}~ a)>0$, then the asymptotic variance of $\ddot V^{\pi}$ and $\ddot V^{\pi}_{SN}$ is strictly larger than that of $\hat V^{\pi}$ and $\hat V^{\pi}_{SN}$.
	\end{enumerate}
\end{theorem}

\noindent The condition in Part 1 of Theorem \ref{prop:true_pscore:stationary} is about the dominating term in the difference between $\hat V^\pi$ and $\tilde V^\pi$.
The proofs of Theorems \ref{prop:estimator:stationary} and \ref{prop:true_pscore:stationary} show that the asymptotic variance of $\hat V^{\pi}$ is the asymptotic variance of
$
	\tilde{V}^\pi-\frac{1}{T}\sum_{t=1}^T \Bigl[\sum_{a=0}^m \mathbb{E}[Y(a)|X_t]\frac{\pi(a|X_t)}{p_{0a}(X_t)}D_{ta}-\theta(X_t)\Bigr].
$
Part 1 of Theorem \ref{prop:true_pscore:stationary} requires that the second term be not always zero so that the asymptotic variance of $\hat V^{\pi}$ is different from that of $\tilde V^{\pi}$. As long as the two variances are not the same, $\hat V^{\pi}$ achieves variance reduction.

Part 2 of Theorem \ref{prop:true_pscore:stationary} requires that $\mathbb{E}[\frac{1}{p_{ta}}|X_t]-\frac{1}{\mathbb{E}[p_{ta}|X_t]}(=\mathbb{E}[\frac{1}{p_{ta}}|X_t]-\frac{1}{p_{0a}(X_t)})\neq 0$ with a positive probability.
This means that $p_t$ is not always the same as the true propensity score $p_{0}(X_t)$, i.e., $F(\cdot|X_t)$ is not degenerate (recall that $p_t$ is drawn from $F(\cdot|X_t)$ whose expected value is $p_0(X_t)$).
Under this condition, $\ddot{V}^\pi$ has a strictly larger asymptotic variance than $\tilde{V}^\pi$ and $\hat{V}^\pi$.

Theorems \ref{prop:bound:stationary} and \ref{prop:true_pscore:stationary} suggest that we should use an estimated score regardless of whether the propensity score is known. To develop some intuition for this result, consider a simple situation where the context $X_t$ always takes some constant value $x$.
Suppose that we are interested in estimation of the expected reward from action $a$, $\mathbb{E}[Y(a)]$.
Since $X$ is constant across rounds, a natural nonparametric estimator for $p_{0a}(x)$ is the proportion of rounds in which action $a$ was chosen: $\hat p_{a}(x)=\frac{\sum_{t=1}^TD_{ta}}{T}$.
The estimator using the estimated propensity score is
$$
\hat V^{\pi}=\frac{1}{T}\sum_{t=1}^TY_t\frac{D_{ta}}{\hat p_{a}(x)}=\frac{1}{\sum_{t=1}^TD_{ta}}\sum_{t=1}^T Y_t D_{ta}.
$$
The estimator using the true propensity score is
$$
\tilde V^{\pi}=\frac{1}{T}\sum_{t=1}^TY_t\frac{D_{ta}}{p_{0a}(x)}=\frac{1}{Tp_{0a}(x)}\sum_{t=1}^T Y_t D_{ta}.
$$
When action $a$ happens to be chosen frequently in a sample so that $\sum_{t=1}^TD_{ta}$ is larger, the absolute value of $\sum_{t=1}^T Y_t D_{ta}$ tends to be larger in the sample.
Because of this positive correlation between $\sum_{t=1}^TD_{ta}$ and the absolute value of $\sum_{t=1}^T Y_t D_{ta}$, $\hat V^\pi$ has a smaller variance than $\tilde V^\pi$, which produces no correlation between the numerator and the denominator.
Similar intuition applies to the comparison between $\ddot V^{\pi}$ and $\hat V^{\pi}$. 


\subsection{How to Estimate Propensity Scores?}\label{section:pscore}
There are several options for the first step estimation of the propensity score.
\begin{enumerate}
	\item A sieve Least Squares (LS) estimator:
	$$
	\hat p_a(\cdot)=\argmin_{p_a(\cdot)\in {\cal H}_{aT}}\frac{1}{T}\sum_{t=1}^T(D_{ta}-p_a(X_t))^2,
	$$
	where ${\cal H}_{aT}=\{p_a(x)=\sum_{j=1}^{k_{aT}}q_{aj}(x)\lambda_{aj}=q^{k_{aT}}(x)'\lambda_a\}$ and $k_{aT}\rightarrow\infty$ as $T\rightarrow\infty$. 
	Here $\{q_{aj}\}_{j=1}^\infty$ is some known basis functions defined on $\cal X$ and $q^{k_{aT}}(\cdot)=(q_{a1}(\cdot),...,q_{ak_{aT}}(\cdot))'$.
	\item A sieve Logit Maximum Likelihood estimator:
	\begin{align*} 
		\hat p(\cdot)&= \argmax_{p(\cdot)\in {\cal H}_{T}}\frac{1}{T}\sum_{t=1}^T \sum_{a=0}^m D_{at}\log p_{a}(X_t),
	\end{align*}
	where
	${\cal H}_T=\{p:{\cal X}\rightarrow (0,1)^{m+1}: p_a(x)=\frac{\exp(R^{k_T}(x)'\lambda_a)}{1+\sum_{a=1}^m\exp(R^{k_T}(x)'\lambda_a)}~\text{for}~ a=1,...,m,  p_0(x)=1-\sum_{a=1}^m p_a(x)\}$.
	Here $R^{k_T}(\cdot)=(R_1(\cdot),...,R_{k_T}(\cdot))'$ and $\{R_j\}_{j=1}^\infty$ is the set of some basis functions.
	\item Prediction of $D_{ta}$ by $X_t$ using a modern machine learning algorithm like random forest, ridge logistic regression, and gradient boosting. 
\end{enumerate}

The above estimators are known to satisfy consistency with a convergence rate faster than $n^{1/4}$ under regularity conditions \citep{Newey1997,cattaneo2010efficient,knight2000asymptotics,blanchard2003rate,buhlmann2011statistics,wager2017estimation}. 

How should one choose a propensity score estimator?
We prefer an estimated score to the true one because it corrects the discrepancy between the realized action assignment in the data and the assignment predicted by the true score.
To achieve this goal, a good propensity score estimator should fit the data better than the true one, which means that the estimator should overfit to some extent.
As a concrete example, in our empirical analysis, random forest produces a larger (worse) variance than gradient boosting and ridge logistic regression (see Figure \ref{figure} and Table \ref{table}).
This is because random forest fits the data worse, which is due to its bagging aspect preventing random forest from overfitting.
In general, however, we do not know which propensity score estimator achieves the best degree of overfitting.
We would therefore suggest that the analyst try different estimators to determine which one is most efficient.


\section{Estimating Asymptotic Variance}\label{section:variance}
We often need to estimate the asymptotic variance of the above estimators. 
For example, variance estimation is crucial for determining whether a counterfactual policy is statistically significantly better than the logging policy. 
We propose an estimator that uses the sample analogue of an expression of the asymptotic variance.
As shown in the proof of Theorem \ref{prop:estimator:stationary}, the asymptotic variance of $\hat V^\pi$ and $\hat V^\pi_{SN}$ is $\mathbb{E}[(g(Y,X,D,V^\pi,p_0)+\alpha(X,D,p_0,\mu_0))^2]$, where $\mu_0:{\cal X}\rightarrow \mathbb{R}^{m+1}$ such that $\mu_0(a|x)=\mathbb{E}[Y(a)|X=x]$ for each $a$ and $x$,
$$
g(Y,X,D,\theta,p)=\sum_{a=0}^m YD_{a}\frac{\pi(a|X)}{p_{a}(X)}-\theta,
$$
and
$$
\alpha(X,D,p,\mu)=-\sum_{a=0}^m\mu(a|X)\frac{\pi(a|X)}{p_{a}(X)}(D_a-p_{a}(X)).
$$

We estimate this asymptotic variance in two steps.
In the first step, we obtain estimates of $V^\pi$ and $p_0$ using the method in Section \ref{estimation}.
In addition, we estimate $\mu_0(a|x)$ by nonparametric regression of $Y_t$ on $X_t$ using the subsample with $D_{ta}=1$ for each $a$.
Denote the estimate by $\hat \mu$.
For this regression, one may use a sieve Least Squares estimator and machine learning algorithms.
In the empirical application below, we use ridge logistic regression.

In the second step, we plug the estimates of $V^\pi$, $p_0$ and $\mu_0$ into the sample analogue of $\mathbb{E}[(g(Y,X,D,V^\pi,p_0)+\alpha(X,D,p_0,\mu_0))^2]$ to estimate the asymptotic variance: when we use $\hat V^\pi$:
\begin{align*}
&\widehat {AVar}(\hat V^\pi)\\
 =& \frac{1}{T}\sum_{t=1}^T(g(Y_t,X_t,D_t,\hat V^\pi,\hat p)+\alpha(X_t,D_t,\hat p,\hat \mu))^2.
\end{align*}
When we use $\hat V^\pi_{SN}$, its asymptotic variance estimator is obtained by replacing $\hat V^\pi$ with $\hat V^\pi_{SN}$ in the above expression.

This asymptotic variance estimator is a two-step generalized-method-of-moment estimator, and is shown to be a consistent estimator under the condition that the first step estimator of $(V^\pi, p_0,\mu_0)$ is consistent and some regularity conditions \cite{Newey1994}.

It is easier to estimate the asymptotic variance of $\tilde V^\pi$ and $\tilde V^\pi_{SN}$ with the true propensity score. 
Their asymptotic variance is $\mathbb{E}[g(Y,X,D,V^\pi,p_0)^2]$ by the standard central limit theorem.
When we use $\tilde V^\pi$, we estimate this asymptotic variance by
$$
\widehat {AVar}(\tilde V^\pi) = \frac{1}{T}\sum_{t=1}^Tg(Y_t,X_t,D_t,\tilde V^\pi,p_0)^2
$$
When we use $\tilde V^\pi_{SN}$, its asymptotic variance estimator is obtained by replacing $\tilde V^\pi$ with $\tilde V^\pi_{SN}$ in the above expression.

\section{Real-World Application}

\begin{table*}[htb]\centering
	\begin{threeparttable}
		\begin{tabular}{l |cc | cc}\hline
			&\multicolumn{2}{c|}{Existing Logging Policy} & \multicolumn{2}{c}{Policy of Choosing Best Action by Context}\\[0.2cm]
			Propensity Score Estimator& CI for Expected Reward & Shrinkage in CI & CI & Shrinkage in CI \\ \hline
			&&&& \\[-0.3em]
			True Score (Benchmark) & $1.036\pm 0.083$& $0$ & $1.140\pm 0.171$ & $0$ \\[0.2cm] 
			Gradient Boosting Machine & $1.047\pm 0.064$& $-22.5\%$ & $1.197\pm 0.131$ & $-23.4\%$\\[0.2cm]
			Ridge Logistic Regression & $0.987\pm 0.058$& $-30.2\%$ & $1.108\pm 0.113$& $-34.1\%$\\[0.2cm]
			Random Forest & $1.036\pm 0.077$& $-6.62\%$ & $1.194\pm 0.159$ & $-7.41\%$\\[0.2cm] \hline
			Sample Size & \multicolumn{4}{c}{$57,619$}\\\hline
		\end{tabular}
		\caption{Improving Ad Design with Lower Uncertainty}\label{table}\par
		~\par
		\fontsize{9.0pt}{10.0pt}\selectfont \textit{Notes}: The first and third columns of this table show 95\% confidence intervals of the expected CTRs $V^{\pi}$ of the logging policy and a hypothetical policy of choosing the best action (ad) for each context. 
			CTRs are multiplied by a constant for confidentiality reasons. 
			We obtain the CTR estimates by the self-normalized inverse probability weighting estimator $\tilde V^\pi_{SN}$ using the true propensity score \citep{Swaminathan2015b} or the estimated propensity score ($\hat V^{\pi}_{SN}$ in Section \ref{estimation}). 
			We estimate standard errors and confidence intervals based on the method described in Section \ref{section:variance}. 
			The second and fourth columns show the size of reductions in confidence interval length, i.e., value $\alpha$ such that the length of the confidence interval is equal to $100-\alpha\%$ of the length of the confidence interval using the true propensity score.\\
	\end{threeparttable}
\end{table*}

We apply our estimators described in Sections \ref{estimation} and \ref{section:variance} to empirically evaluate and optimize the design of online advertisements. 
This application uses proprietary data provided by CyberAgent Inc., which we described in the introduction. 
This company uses a contextual bandit algorithm to determine the visual design of advertisements assigned to user impressions (there are four design choices). 
This algorithm produces logged bandit data. 
We use this logged bandit data and our estimators to improve their advertisement design for maximizing the click through rates (CTR). 
In the notation of our theoretical framework, reward $Y$ is a click, action $a$ is one of the four possible individual advertisement designs, and context $X$ is user and ad characteristics used by the company's logging policy. 

The logging policy (the company's existing contextual bandit algorithm) works as follows. 
For each round, the logging policy first randomly samples each action's predicted reward from a beta distribution. 
This beta distribution is parametrized by the predicted CTR for each context, where the CTR prediction is based on a Factorization Machine \citep{rendle2010factorization}. 
The logging policy then chooses the action (advertisement) with the largest sampled reward prediction. 
The logging policy and the underlying CTR prediction stay the same for all rounds in each day. 
Each day therefore performs the role of a batch in the model in Section 2. 
This somewhat nonstandard logging policy and the resulting log data are an example of our DGP in Section \ref{model}. 

This logging policy may have room for improvement for several reasons. 
First, the logging policy randomly samples advertisements and does not necessarily choose the advertisement with the best predicted CTR. 
Also, the logging policy uses a predictive Factorization Machine for its CTR prediction, which may be different from the causal CTR (the causal effect of each advertisement on the probability of a click). 

To improve on the logging policy, we first estimate the propensity score by random forest, ridge logistic regression, or gradient boosting (implemented by XGBoost). 
These estimators are known to satisfy the regularity conditions (e.g. consistency) required for our theoretical results, as explained in Section \ref{section:pscore}. 

With the estimated propensity score, we then use our estimator $\hat V^{\pi}_{SN}$ to estimate the expected reward from two possible policies: 
(1) the logging policy and (2) a counterfactual policy that chooses the best action (advertisement) that is predicted to maximize the CTR conditional on each context. 
To implement this counterfactual policy, we estimate $\mathbb{E}[Y(a)|X]$ by ridge logistic regression for each action $a$ and context $X$ used by the logging policy (we apply one-hot encoding to categorical variables in $X$). 
Given each context $X$, the counterfactual policy then chooses the action with the highest estimated value of $\mathbb{E}[Y(a)|X]$.

Importantly, we use separate data sets for the two estimation tasks (one for the best actions and the other for the expected reward from the hypothetical policy). 
Specifically, we use data logged during April 20-26, 2018 for estimating the best actions and data during April 27-29 for estimating the expected reward. 
This data separation allows us to avoid overfitting and overestimation of the CTR gains from the counterfactual policy. 

As a benchmark, we also estimate the same expected rewards based on \citet{Swaminathan2015b}'s self-normalized estimator $\tilde V^\pi_{SN}$, which uses the true propensity score. 
The resulting estimates show the following result: 

\begin{quote}
	\textbf{Empirical Result A.} Consistent with Theorems \ref{prop:estimator:stationary}-\ref{prop:true_pscore:stationary}, our estimator $\hat V^{\pi}_{SN}$ with the estimated score is statistically more efficient than the benchmark $\tilde V^\pi_{SN}$ with the true score. 
\end{quote}

This result is reported in Figure \ref{figure} and Table \ref{table}, where the confidence intervals about the predicted CTR using ``True Propensity Score (Benchmark)" are less precise (wider) than those using estimated propensity scores (regardless of which one of the three score estimators to use). 
The magnitude of this shrinkage in the confidence intervals and standard errors is 6-34\%, depending on how to estimate the propensity score. 

This variance reduction allows us to conclude that the logging policy is below the lower bound of the confidence interval of the hypothetical policy, giving us confidence in the following implication: 

\begin{quote}
	\textbf{Empirical Result B.} Compared to the logging policy, the hypothetical policy (choosing the best advertisement given each context) improves the CTR by 10-15\% statistically significantly at the 5\% significance level. 
\end{quote}

\section{Conclusion} 

We have investigated the most statistically efficient use of batch bandit data for estimating the expected reward from a counterfactual policy. 
Our estimators minimize the asymptotic variance among all asymptotically normal estimators (Theorem \ref{prop:estimator:stationary}). 
By contrast, standard estimators have larger asymptotic variances (Theorem \ref{prop:true_pscore:stationary}). 

We have also applied our estimators to improve online advertisement design. 
Compared to the frontier benchmark $\tilde V^\pi_{SN}$, our reward estimator $\hat V^{\pi}_{SN}$ provides the company with more statistical confidence in how to improve on its existing bandit algorithm (Empirical Results A and B). 
The hypothetical policy of choosing the best advertisement given user characteristics would improve the click through rate by 10-15\% at the 5\% significance level. 
These empirical results thus highlight the practical values of Theorems \ref{prop:estimator:stationary}-\ref{prop:true_pscore:stationary}. \\

\noindent \textbf{Acknowledgments.} 
We are grateful to seminar participants at ICML/IJCAI/AAMAS Workshop ``Machine Learning for Causal Inference, Counterfactual Prediction, and Autonomous Action (CausalML)" and RIKEN Center for Advanced Intelligence Project, especially Junya Honda, Masaaki Imaizumi, Atsushi Iwasaki, Kohei Kawaguchi, and Junpei Komiyama. 


\bibliography{aaai2019-paper}
\bibliographystyle{aaai}

\appendix

\section*{Appendices}

\section{Defining Semiparametric Efficiency Bound}\label{SEB}
We present the definition of semiparametric efficiency bound based on \citet{Bickel1993}.
Let $X_1,...,X_n$ be an i.i.d. sample from the probability distribution $P_0$ on $(\mathbf{X},\cal{B})$, where $\mathbf{X}$ is some Euclidean sample space and $\cal{B}$ is its Borel $\sigma$-field.
Let $\mu$ be a fixed $\sigma$-finite measure on $(\mathbf{X},\cal{B})$, and let $\mathbf{M}$ be the collection of all probability measures dominated by $\mu$.
Consider a subset $\mathbf{P}$ of $\mathbf{M}$ such that $P_0\in \mathbf{P}$, and a parameter $v:\mathbf{P}\rightarrow \mathbb{R}$.

We first define a {\it regular parametric model}.
Consider a subset $\mathbf{Q}$ of $\mathbf{P}$ that has a parametrization $\theta\rightarrow P_\theta$ such that
$$
\mathbf{Q}=\{P_\theta:\theta\in\Theta\},
$$
where $\Theta$ is a subset of $\mathbb{R}^k$.
Let $p(\theta)=\frac{dP_\theta}{d\mu}$, a density of $P_\theta$, and $s(\theta)=\sqrt{p(\theta)}$.
In the following, $L_2(\mu)$ is the Hilbert space of $\mu$-square integrable functions, $|\cdot|$ is the Euclidean norm, and $\Vert\cdot\Vert$ is the Hilbert norm in $L_2(\mu)$: $\Vert f\Vert\equiv \int f^2 d\mu$.

\begin{definition}[Definition 2.1.1 in \citet{Bickel1993}]
	$\theta_0$ is a {\it regular point} of the parametrization $\theta\rightarrow P_\theta$ if $\theta_0$ is an interior point of $\Theta$, and
	\renewcommand\labelenumi{(\roman{enumi})}
	\begin{enumerate}
		\setlength{\itemindent}{10pt}
		\item The map $\theta\rightarrow s(\theta)$ from $\Theta$ to $L_2(\mu)$ is Fr\'echet differentiable at $\theta_0$: there exists a vector $\dot s(\theta_0)=(\dot s_1(\theta_0),...,\dot s_k(\theta_0))'$ such that
		$$
		\Vert s(\theta_0+h)-s(\theta_0)-\dot s(\theta_0)'h\Vert=o(|h|) \ \ \ \text{as $h\rightarrow 0$}.
		$$
		\item The $k\times k$ matrix $\int \dot s(\theta_0)\dot s(\theta_0)' d\mu$ is nonsingular.
	\end{enumerate}
\end{definition}

\begin{definition}[Definition 2.1.2 in \citet{Bickel1993}]
	A parametrization $\theta\rightarrow P_\theta$ is {\it regular} if:
	\renewcommand\labelenumi{(\roman{enumi})}
	\begin{enumerate}
		\setlength{\itemindent}{10pt}
		\item Every point of $\Theta$ is regular.
		\item The map $\theta\rightarrow \dot s_i(\theta)$ is continuous from $\Theta$ to $L_2(\mu)$ for $i=1,...,k$.
	\end{enumerate}
\end{definition}
\noindent
We call $\mathbf{Q}$ a {\it regular parametric model} if it has a regular parametrization.

Now let $q(\theta)=v(P_\theta)$.
Fix $P=P_\theta\in \mathbf{Q}$ and suppose $q$ has a $1\times k$ total differential vector $\dot q$ at $\theta$.
Define
$$
I^{-1}(P|v,\mathbf{Q})=\dot q(\theta)I^{-1}(\theta)\dot q(\theta)',
$$
where
$$
I(\theta) = 4\int \dot s(\theta)\dot s(\theta)'d\mu.
$$
Suppose that there exists a regular parametric model $\mathbf{Q}\subset \mathbf{P}$ that contains $P_0$.

\begin{definition}
	The {\it semiparametric efficiency bound} for $v$ is defined by
	\begin{align*}
		I^{-1}(P_0|v,\mathbf{P}) \equiv&\sup\{I^{-1}(P_0|v,\mathbf{Q}): \mathbf{Q}\subset \mathbf{P} \text{ and $\mathbf{Q}$ is a}\\
		&\text{regular parametric model that contains $P_0$}\}.
	\end{align*}
\end{definition}

\section{Proofs}\label{proofs}
\noindent \textit{Proof of Lemma \ref{prop:bound:stationary}.}
The derivation of the semiparametric efficiency bound follows the approach of \citet{Hahn1998}, \citet{HIR2003}, \citet{chen2008semiparametric}, \citet{cattaneo2010efficient} and \citet{Newey1990}.
The proof proceeds in four steps: (i) characterize the tangent set for all regular parametric submodels, (ii) verify that the parameter of interest is pathwise differentiable, (iii) verify that the efficient influence function lies in the tangent set, and (iv) calculate the expected square of the influence function.

Consider a regular parametric submodel of the joint distribution of $(Y,D,X)$ with parameter $\beta$ and the likelihood given by
$$
f(y,d,x;\beta)=\bigl\{\Pi_{a=0}^m [f_a(y|x;\beta)p_a(x;\beta)]^{d_a} \bigr\} f_X(x;\beta),
$$
where $f_a(y|x;\beta)$ is the conditional density of $Y(a)$ given $X$, $p_a(x;\beta)=\Pr\{D_a=1|X;\beta\}$, and $f_X(x;\beta)$ is the density of $X$.
The log-likelihood function is
\begin{align*}
	&\log f(y,d,x;\beta) \\
	= &\sum_{a=0}^m d_a [\log f_a(y|x;\beta)+\log p_a(x;\beta)] + \log f_X(x;\beta).
\end{align*}

The corresponding score is
\begin{align*}
	S(y,d,x;\beta) 
	&= \frac{d}{d\beta} \log f(y,d,x;\beta) \\
	&= \sum_{a=0}^m d_a [s_a(y|x;\beta)+\frac{\dot p_a(x;\beta)}{p_a(x;\beta)}] + t(x;\beta),
\end{align*}
where $s_a(y|x;\beta)=\frac{d}{d\beta}\log f_a(y|x;\beta)$, $\dot p_a(x;\beta)=\frac{d}{d\beta}  p_a(x;\beta)$, and $t(x;\beta)=\frac{d}{d\beta} \log f_X(x;\beta)$.
The tangent set of this model is therefore given by
$$
{\cal T} =\{\sum_{a=0}^m d_a [s_a(y|x;\beta_0)+\frac{\dot p_a(x;\beta_0)}{p_a(x;\beta_0)}] + t(x;\beta_0)\},
$$
where $\int s_a(y|x;\beta_0)f_a(y|x;\beta_0)dy=0$ for all $x$ and $a$, $\sum_{a=0}^m \dot p_a(x;\beta_0)=0$ for all $x$, and $\int t(x;\beta_0)f_X(x;\beta_0)dx=0$.

Now let $V^\pi(\beta)$ be our parameter of interest as a function of $\beta$:
\begin{align*}
	V^\pi(\beta)&=\mathbb{E}_{\beta}[\sum_{a=0}^m Y(a)\pi(a|X)]\\
	&=\sum_{a=0}^m \int\int y\pi(a|x) f_a(y|x;\beta)f_X(x;\beta)dydx.
\end{align*}
Differentiation of this under the integral gives
\begin{align*}
	&\frac{\partial V^\pi(\beta)}{\partial \beta} \\
	=& \sum_{a=0}^m \int\int y\pi(a|x) \frac{\partial}{\partial \beta}f_a(y|x;\beta)f_X(x;\beta)dydx\\
	&+\sum_{a=0}^m \int\int y\pi(a|x)f_a(y|x;\beta)\frac{\partial}{\partial \beta}f_X(x;\beta)dydx\\
	=& \sum_{a=0}^m \int\int y\pi(a|x) s_a(y|x;\beta) f_a(y|x;\beta)f_X(x;\beta)dydx\\
	&+\sum_{a=0}^m \int\int y\pi(a|x)f_a(y|x;\beta) t(x;\beta)f_X(x;\beta)dydx\\
	=& \sum_{a=0}^m \mathbb{E}_{\beta}[Y(a)\pi(a|X)s_a(Y(a)|X;\beta)]\\
	&+\sum_{a=0}^m \mathbb{E}_{\beta}[\mathbb{E}_{\beta}[Y(a)|X]\pi(a|X)t(X;\beta)]\\
	=& \mathbb{E}_{\beta}\bigl[\sum_{a=0}^m Y(a)\pi(a|X)s_a(Y(a)|X;\beta)+\theta(X;\beta)t(X;\beta)\bigr],
\end{align*}
where $\theta(X;\beta)=\sum_{a=0}^m \mathbb{E}_{\beta}[Y(a)|X]\pi(a|X)$.

$V^\pi$ is pathwise differentiable if there exists a function $\Psi(y,d,x)$ such that $\mathbb{E}[\Psi(Y,D,X)^2]<\infty$ and for all regular parametric submodels
\begin{align}\label{eq:pathwise_dif}
	\frac{\partial V^\pi(\beta_0)}{\partial \beta}=\mathbb{E}[\Psi(Y,D,X)S(Y,D,X;\beta_0)].
\end{align}
Let
\begin{align*}
	&\Psi(Y,D,X)\\
	=&\sum_{a=0}^m(Y-\mathbb{E}[Y(a)|X])D_a\frac{\pi(a|X)}{p_{0a}(X)}+\theta(X)-V^\pi.
\end{align*}
We first verify that $\mathbb{E}[\Psi(Y,D,X)^2]<\infty$.
Since $\mathbb{E}[(Y-\mathbb{E}[Y(a)|X])D_a\frac{\pi(a|X)}{p_{0a}(X)}|X]=0$ by the independence of $D$ and $Y(\cdot)$ conditional on $X$, we have that $\mathbb{E}[(Y-\mathbb{E}[Y(a)|X])D_a\frac{\pi(a|X)}{p_{0a}(X)}(\theta(X)-V^{\pi})]=0$ for all $a$. Also, $\mathbb{E}[(Y-\mathbb{E}[Y(a)|X])D_a\frac{\pi(a|X)}{p_{0a}(X)} (Y-\mathbb{E}[Y(a')|X])D_{a'}\frac{\pi(a'|X)}{p_{0a'}(X)}]=0$ because $D_aD_{a'}=0$ for any $a\neq a'$.
It then follows that
\begin{align*}
	&\mathbb{E}[\Psi(Y,D,X)^2]\\
	=& \mathbb{E}[\sum_{a=0}^m (Y-\mathbb{E}[Y(a)|X])^2D_a\frac{\pi(a|X)^2}{p_{0a}(X)^2}+(\theta(X)-V^{\pi})^2]\\
	=&\mathbb{E}[\sum_{a=0}^m \mathbb{V}[Y(a)|X]\frac{\pi(a|X)^2}{p_{0a}(X)}+(\theta(X)-V^{\pi})^2],
\end{align*}
where for the last equality, we use the independence of $D$ and $Y(\cdot)$ conditional on $X$.
Under Assumptions \ref{nondegeneracy} and \ref{finite_variance}, $\mathbb{E}[\Psi(Y,D,X)^2]<\infty$.

We next verify that equation (\ref{eq:pathwise_dif}) holds.
The RHS is
\begin{align*}
	&\mathbb{E}\Bigl[\bigl\{\sum_{a=0}^m(Y-\mathbb{E}[Y(a)|X])D_a\frac{\pi(a|X)}{p_{0a}(X)}+\theta(X)-V^\pi\bigr\}\\
	&\times \bigl\{\sum_{a=0}^m D_a [s_a(Y|X;\beta_0)+\frac{\dot p_a(X;\beta_0)}{p_a(X;\beta_0)}] + t(X;\beta_0)\bigr\}\Bigr]\\
	=&\mathbb{E}\Bigl[\sum_{a=0}^m(Y-\mathbb{E}[Y(a)|X])D_a\frac{\pi(a|X)}{p_{0a}(X)}s_a(Y|X;\beta_0)\\
	+&\sum_{a=0}^m(Y-\mathbb{E}[Y(a)|X])D_a\frac{\pi(a|X)}{p_{0a}(X)}[\frac{\dot p_a(X;\beta_0)}{p_a(X;\beta_0)}+t(X;\beta_0)]\\
	+&\theta(X) t(X;\beta_0)\Bigr]\\
	=&\mathbb{E}\Bigl[\sum_{a=0}^mY(a)\pi(a|X)s_a(Y(a)|X;\beta_0)+\theta(X) t(X;\beta_0)\Bigr]\\
	=&\frac{\partial V^\pi(\beta_0)}{\partial \beta},
\end{align*}
where the first equality holds because $D_aD_{a'}=0$ for $a\neq a'$, $\mathbb{E}[\sum_{a=0}^m D_a [s_a(Y|X;\beta_0)+\frac{\dot p_a(X;\beta_0)}{p_a(X;\beta_0)}]|X]=\sum_{a=0}^m \{p_{0a}(X)\mathbb{E}[s_a(Y(a)|X;\beta_0)|X]+\dot p_a(X;\beta_0)\}=0$, and $\mathbb{E}[V^\pi S(Y,D,X;\beta_0)]=0$, the second equality holds because $\mathbb{E}[D_a s_a(Y|X;\beta_0)|X]=p_{0a}(X)\mathbb{E}[s_a(Y(a)|X;\beta_0)|X]=0$, and $\mathbb{E}[(Y-\mathbb{E}[Y(a)|X])D_a|X]=(\mathbb{E}[Y(a)|X]-\mathbb{E}[Y(a)|X])p_{0a}(X)=0$, and the last equality holds because $\mathbb{E}_{\beta_0}[\cdot]=\mathbb{E}[\cdot]$ and $\theta(X;\beta_0)=\theta(X)$.
We therefore conclude that $V^\pi$ is pathwise differentiable.

Finally, we can verify that $\Psi\in \cal{T}$, since $\Psi$ is the score of a parametric submodel such that $s_a(y|x;\beta_0)= (y-\mathbb{E}[Y(a)|X=x])\frac{\pi(a|x)}{p_{0a}(x)}$, $\dot p_a(x;\beta_0)=0$, and $t(x;\beta_0)=\theta(x)-V^\pi$ for all $x$ and $a$.

By Theorem 3.1 of \citet{Newey1990}, the semiparametric efficiency bound is the expected square of the projection of $\Psi$ on $\cal{T}$.
Since $\Psi\in \cal{T}$, the projection on $\cal{T}$ is itself, and the semiparametric efficiency bound is $\mathbb{E}[\Psi(Y,D,X)^2]$.
\qed
\par
\bigskip
\noindent
\noindent \textit{Proof of Theorem \ref{prop:estimator:stationary}.}
We first show that $\hat V^\pi$ achieves the semiparametric efficiency bound.
Let $p(\cdot)=(p_0(\cdot),...,p_m(\cdot))'$ denote a candidate propensity vector, and let $\tilde p(\cdot)=(p_1(\cdot),...,p_m(\cdot))'$.
Let $g(\cdot)$ and $\rho(\cdot)$ be the following scalar valued function and $m\times 1$ vector valued function:
\begin{align*}
	&g(Z,\theta,\tilde p)\\
	=&YD_{0}\frac{\pi(0|X)}{1-\sum_{a=1}^m p_{a}(X)}+\sum_{a=1}^m YD_{a}\frac{\pi(a|X)}{p_{a}(X)}-\theta,
\end{align*}
and
$$
\rho(Z,\tilde p(X))=\left(\begin{array}{c}
D_1-p_1(X)\\
\vdots \\
D_m-p_m(X)
\end{array}\right),
$$
where $Z=(Y,X,D)$.
Then, $V^{\pi}$ is identified by the unconditional moment restriction
\begin{align}\label{eq:moment_g}
	\mathbb{E}[g(Z,V^{\pi},\tilde p_0)]=0,
\end{align}
and $\tilde p_0(\cdot)$ is identified by the conditional moment restriction
\begin{align}\label{eq:moment_rho}
	\mathbb{E}[\rho(Z,\tilde p_0(X))|X]=0~\text{a.s.}~X.
\end{align}
In addition, $\hat V^{\pi}$ is characterized by the solution to
$$
\frac{1}{T}\sum_{t=1}^T g(Z_t,\hat V^{\pi},\hat {\tilde p})=0,
$$
where $\hat {\tilde p}(\cdot)=(\hat p_1(\cdot),...,\hat p_m(\cdot))$ is a nonparametric consistent estimator.
We have that
$$
\sqrt{T}(\hat V^{\pi}-V^{\pi})=\frac{1}{\sqrt{T}}\sum_{t=1}^T g(Z_t,V^{\pi},\hat {\tilde p}).
$$

We use \citet{Ackerberg2014}'s results to show that the asymptotic variance of $\sqrt{T}(\hat V^{\pi}-V^{\pi})$ is equal to the semiparametric efficiency bound given by Lemma \ref{prop:bound:stationary}.
Let
$$
D(X,V^{\pi},\tilde p_0)\equiv\frac{\partial \mathbb{E}[g(Z,V^{\pi},\tilde p_0)|X]}{\partial \tilde p'}.
$$
Here, the $a$-th element of $D(X,\theta_0,\tilde p_0)$ is given by
\begin{align*}
	&D_a(X,V^{\pi},\tilde p_0)\\
	=&\mathbb{E}[YD_{0}|X]\frac{\pi(0|X)}{p_{00}(X)^2}-\mathbb{E}[YD_{a}|X]\frac{\pi(a|X)}{p_{0a}(X)^2}\\
	=&\mathbb{E}[Y(0)|X]\frac{\pi(0|X)}{p_{00}(X)}-\mathbb{E}[Y(a)|X]\frac{\pi(a|X)}{p_{0a}(X)}
\end{align*}
for $a=1,...,m$, where $p_{00}(X)=1-\sum_{a=1}^m p_{0a}(X)$, and the second equality follows from the fact that $D$ and $Y(\cdot)$ are independent conditional on $X$.
Let $m(X,\tilde p(X))\equiv \mathbb{E}[\rho(Z,\tilde p(X))|X]=\mathbb{E}[\tilde D|X]-\tilde p(X)$, where $\tilde D=(D_1,...,D_m)'$.
We have that $\frac{\partial m(X,\tilde p_0(X))}{\partial \tilde p'}=-I_m$, where $I_m$ is an $m\times m$ identity matrix.

Condition 1.(i)-(iii) of \citet{Ackerberg2014} trivially hold.
To see that Condition 1.(iv) holds, note that
\begin{align*}
	&\frac{\partial \mathbb{E}[g(Z,V^{\pi},\tilde p_0)]}{\partial \tilde p}[v]\\
	=&\mathbb{E}[D(X,V^{\pi},\tilde p_0)v(X)]\\
	=&\mathbb{E}[\sum_{a=1}^m(\mathbb{E}[Y(0)|X]\frac{\pi(0|X)}{p_{00}(X)}-\mathbb{E}[Y(a)|X]\frac{\pi(a|X)}{p_{0a}(X)})v_a(X)]
\end{align*}
for $v\in {\cal V}$, where ${\cal V}$ is a linear subspace of the space of $m$-dimensional square integrable functions with respect to $X$.
$\frac{\partial \mathbb{E}[g(Z,V^{\pi},\tilde p_0)]}{\partial \tilde p}[\cdot]$ is a linear functional on ${\cal V}$.
We have
\begin{align*}
	&\sup_{v\neq 0, v\in {\cal V}}\frac{\left|\frac{\partial \mathbb{E}[g(Z,V^{\pi},\tilde p_0)]}{\partial \tilde p}[v]\right|^2}{\mathbb{E}[v(X)'v(X)]}\\
	=&\sup_{v\neq 0, v\in {\cal V}}\left|\frac{\partial \mathbb{E}[g(Z,V^{\pi},\tilde p_0)]}{\partial \tilde p}[\frac{v}{\mathbb{E}[v(X)'v(X)]^{1/2}}]\right|^2\\
	=&\sup_{v\in \bar{{\cal V}}}\left|\frac{\partial \mathbb{E}[g(Z,V^{\pi},\tilde p_0)]}{\partial \tilde p}[v]\right|^2\\
	=&\sup_{v\in \bar{{\cal V}}}\left|\sum_{a=1}^m\mathbb{E}[D_a(X,V^\pi,\tilde p_0)v_a(X)]\right|^2\\
	\le&\sup_{v\in \bar{{\cal V}}}\sum_{a=1}^m\sum_{a'=1}^m|\mathbb{E}[D_a(X,V^\pi,\tilde p_0)v_a(X)]|\\
	&\hspace{5em}\times|\mathbb{E}[D_{a'}(X,V^\pi,\tilde p_0)v_{a'}(X)]|\\
	\le&\sup_{v\in \bar{{\cal V}}}\sum_{a=1}^m\sum_{a'=1}^m\mathbb{E}[D_a(X,V^\pi,\tilde p_0)^2]^{1/2}\mathbb{E}[v_a(X)^2]^{1/2}\\
	&\hspace{5em}\times \mathbb{E}[D_{a'}(X,V^\pi,\tilde p_0)^2]^{1/2}\mathbb{E}[v_{a'}(X)^2]^{1/2}\\
	\le&\sum_{a=1}^m\sum_{a'=1}^m\mathbb{E}[D_a(X,V^\pi,\tilde p_0)^2]^{1/2}\mathbb{E}[D_{a'}(X,V^\pi,\tilde p_0)^2]^{1/2}
\end{align*}
where $\bar{{\cal V}}=\{v'=v/\mathbb{E}[v(X)'v(X)]^{1/2}: v\neq 0, v\in {\cal V}\}=\{v\in {\cal V}: \mathbb{E}[v(X)'v(X)]=1\}$, the first equality uses the linearity of $\frac{\partial \mathbb{E}[g(Z,V^{\pi},\tilde p_0)]}{\partial \tilde p}[\cdot]$, the second equality uses the definition of $\bar{{\cal V}}$, the first inequality uses the triangle inequality, the second inequality uses the Cauchy-Schwarz inequality, and the last inequality holds because $\mathbb{E}[v_a(X)^2]\le \mathbb{E}[v(X)'v(X)]=1$ for all $a$ and all $v\in \bar{{\cal V}}$.
Under Assumptions \ref{nondegeneracy} and \ref{finite_variance}, $\mathbb{E}[D_a(X,V^\pi,\tilde p_0)^2]<\infty$ for all $a$, and thus $\frac{\partial \mathbb{E}[g(Z,V^{\pi},\tilde p_0)]}{\partial \tilde p}[\cdot]$ is a bounded linear functional.


Now note that the function $g$ depends on $\tilde p(\cdot)$ only through $\tilde p(X)$.
By Lemma 2 and Proposition 1 of \citet{Ackerberg2014}, the asymptotic variance of $\sqrt{T}(\hat V^{\pi}-V^{\pi})$ is equal to $\mathbb{V}[\tilde g(Z,V^{\pi},\tilde p_0)]$, where
\begin{align*}
	&\tilde g(Z,V^{\pi},\tilde p_0)\\
	\equiv& g(Z,V^{\pi},\tilde p_0)-D(X,V^{\pi},\tilde p_0)\\
	&\hspace{6em}\bigl(\frac{\partial m(X,\tilde p_0(X))}{\partial \tilde p'}\bigr)^{-1}\rho(Z,\tilde p_0(X))\\
	=& \sum_{a=0}^m YD_{a}\frac{\pi(a|X)}{p_{0a}(X)}-V^{\pi}+\sum_{a=1}^m\{\mathbb{E}[Y(0)|X]\frac{\pi(0|X)}{p_{00}(X)}\\
	&\hspace{5em}-\mathbb{E}[Y(a)|X]\frac{\pi(a|X)}{p_{0a}(X)}\}(D_a-p_{0a}(X))\\
	=& \sum_{a=0}^m YD_{a}\frac{\pi(a|X)}{p_{0a}(X)}-V^{\pi}\\
	&+\mathbb{E}[Y(0)|X]\frac{\pi(0|X)}{p_{00}(X)}\sum_{a=1}^m(D_a-p_{0a}(X))\\
	&-\sum_{a=1}^m\mathbb{E}[Y(a)|X]\frac{\pi(a|X)}{p_{0a}(X)}(D_a-p_{0a}(X))\\
	=& \sum_{a=0}^m YD_{a}\frac{\pi(a|X)}{p_{0a}(X)}-V^{\pi}\\
	&+\mathbb{E}[Y(0)|X]\frac{\pi(0|X)}{p_{00}(X)}(1-D_0-(1-p_{00}(X)))\\
	&-\sum_{a=1}^m\mathbb{E}[Y(a)|X]\frac{\pi(a|X)}{p_{0a}(X)}(D_a-p_{0a}(X))\\
	=& \sum_{a=0}^m YD_{a}\frac{\pi(a|X)}{p_{0a}(X)}-V^{\pi}\\
	&-\sum_{a=0}^m\mathbb{E}[Y(a)|X]\frac{\pi(a|X)}{p_{0a}(X)}(D_a-p_{0a}(X))\\
	=& \sum_{a=0}^m (Y-\mathbb{E}[Y(a)|X])D_a\frac{\pi(a|X)}{p_{0a}(X)}\\
	&+\sum_{a=0}^m\mathbb{E}[Y(a)|X]\pi(a|X)-V^{\pi}\\
	=& \sum_{a=0}^m (Y-\mathbb{E}[Y(a)|X])D_a\frac{\pi(a|X)}{p_{0a}(X)}+\theta(X)-V^{\pi}.
\end{align*}
By the moment restrictions (\ref{eq:moment_g}) and (\ref{eq:moment_rho}), $\mathbb{E}[\tilde g(Z,V^{\pi},\tilde p_0)]=0$.
It then follows that
\begin{align*}
	&\mathbb{V}[\tilde g(Z,V^{\pi},\tilde p_0)]\\
	=&\mathbb{E}[\tilde g(Z,V^{\pi},\tilde p_0)^2]\\
	=&\mathbb{E}[\sum_{a=0}^m \mathbb{V}[Y(a)|X]\frac{\pi(a|X)^2}{p_{0a}(X)}+(\theta(X)-V^{\pi})^2],
\end{align*}
where we have shown the last equality in the proof of Lemma \ref{prop:bound:stationary}.

To show that $\hat V^\pi_{SN}$ has the same asymptotic variance as $\hat V^\pi$, it suffices to show that the denominator of $\hat V^\pi_{SN}$, $\frac{1}{T}\sum_{t=1}^T\sum_{a=0}^mD_{ta}\frac{\pi(a|X_t)}{\hat p_a(X_t)}$, converges in probability to one.
Denote this denominator by $\hat\beta$.
Let $h(\cdot)$ be the following scalar valued function:
$$
h(Z,\beta,p)=\sum_{a=0}^mD_{a}\frac{\pi(a|X)}{p_a(X)}-\beta.
$$
Also, let $\beta_0$ be the solution to the following moment condition:
$$
\mathbb{E}[h(Z,\beta,p_0)]=0,
$$
i.e., $\beta_0=\mathbb{E}[\sum_{a=0}^mD_{a}\frac{\pi(a|X)}{p_{0a}(X)}]$.
$\hat\beta$ is characterized by the solution to
$$
\frac{1}{T}\sum_{t=1}^Th(Z_t,\beta,\hat p)=0.
$$
$\hat \beta$ is a semiparametric two-step GMM estimator, and it is shown that $\hat \beta$ is consistent for $\beta_0$ under the condition that $\hat p(\cdot)$ is a consistent estimator for $p_0(\cdot)$ and regularity conditions \cite{Newey1994}.
Since $\beta_0=\mathbb{E}[\sum_{a=0}^mD_{a}\frac{\pi(a|X)}{p_{0a}(X)}]=\mathbb{E}[\sum_{a=0}^m\pi(a|X)]=1$, $\hat \beta$ converges in probability to one.
\qed
\par
\noindent
\noindent \textit{Proof of Theorem \ref{prop:true_pscore:stationary}.}
We use the same notations as those in the proof of Theorem \ref{prop:estimator:stationary}.
We first compare $\tilde V^\pi$ to $\hat V^\pi$.
We have that
$$
\sqrt{T}(\tilde V^{\pi}-V^{\pi})=\frac{1}{\sqrt{T}}\sum_{t=1}^T g(Z_t,V^{\pi},\tilde p_0).
$$
By the central limit theorem, the asymptotic variance of $\sqrt{T}(\tilde V^{\pi}-V^{\pi})$ is $\mathbb{V}[g(Z,V^{\pi},\tilde p_0)]$.
We compare this to $\mathbb{V}[\tilde g(Z,V^{\pi},\tilde p_0)]$.
From the proof of Theorem \ref{prop:estimator:stationary}, we can write:
\begin{align*}
	&\tilde g(Z,V^{\pi},\tilde p_0)\\
	=&g(Z,V^{\pi},\tilde p_0)-\sum_{a=0}^m\mathbb{E}[Y(a)|X]\frac{\pi(a|X)}{p_{0a}(X)}(D_a-p_{0a}(X)).
\end{align*}
It follows that
\begin{align*}
	&\mathbb{V}[g(Z,V^{\pi},\tilde p_0)]-\mathbb{V}[\tilde g(Z,V^{\pi},\tilde p_0)]\\
	=&\mathbb{E}[g(Z,V^{\pi},\tilde p_0)^2]-\mathbb{E}[\tilde g(Z,V^{\pi},\tilde p_0)^2]\\
	=&2\mathbb{E}\bigl[g(Z,V^{\pi},\tilde p_0)\sum_{a=0}^m\mathbb{E}[Y(a)|X]\frac{\pi(a|X)}{p_{0a}(X)}(D_a-p_{0a}(X))\bigr]\\
	&-\mathbb{E}\Bigl[\bigl(\sum_{a=0}^m\mathbb{E}[Y(a)|X]\frac{\pi(a|X)}{p_{0a}(X)}(D_a-p_{0a}(X))\bigr)^2\Bigr].
\end{align*}
The first term equals
\begin{align*}
	&2\mathbb{E}\bigl[(\sum_{a=0}^m YD_{a}\frac{\pi(a|X)}{p_{0a}(X)}-V^{\pi})\\
	&\hspace{3em}\sum_{a=0}^m\mathbb{E}[Y(a)|X]\frac{\pi(a|X)}{p_{0a}(X)}(D_a-p_{0a}(X))\bigr]\\
	=&2\mathbb{E}\bigl[(\sum_{a=0}^m YD_{a}\frac{\pi(a|X)}{p_{0a}(X)})\\
	&\hspace{3em}\sum_{a=0}^m\mathbb{E}[Y(a)|X]\frac{\pi(a|X)}{p_{0a}(X)}(D_a-p_{0a}(X))\bigr]\\
	=&2\mathbb{E}\bigl[\sum_{a=0}^m YD_{a}\frac{\pi(a|X)}{p_{0a}(X)}\mathbb{E}[Y(a)|X]\frac{\pi(a|X)}{p_{0a}(X)}(D_a-p_{0a}(X))\\
	&\hspace{3em}+\sum_{a=0}^m\sum_{a'\neq a}YD_{a}\frac{\pi(a|X)}{p_{0a}(X)}\mathbb{E}[Y(a')|X]\\
	&\hspace{8em}\frac{\pi(a'|X)}{p_{0a'}(X)}(D_{a'}-p_{0a'}(X))\bigr]\\
	=&2\mathbb{E}\bigl[\sum_{a=0}^m YD_{a}(1-p_{0a}(X))\mathbb{E}[Y(a)|X]\frac{\pi(a|X)^2}{p_{0a}(X)^2}
	\\
	&-\sum_{a=0}^m\sum_{a'\neq a}YD_{a}p_{0a'}(X)\mathbb{E}[Y(a')|X]\frac{\pi(a|X)\pi(a'|X)}{p_{0a}(X)p_{0a'}(X)}\bigr]\\
	=&2\mathbb{E}\bigl[\sum_{a=0}^m (1-p_{0a}(X))\mathbb{E}[Y(a)|X]^2\frac{\pi(a|X)^2}{p_{0a}(X)}
	\\
	&-\sum_{a=0}^m\sum_{a'\neq a}\mathbb{E}[Y(a)|X]\mathbb{E}[Y(a')|X]\pi(a|X)\pi(a'|X)\bigr].
\end{align*}
The second term equals
\begin{align*}
	&-\mathbb{E}\bigl[\sum_{a=0}^m\mathbb{E}[Y(a)|X]^2\frac{\pi(a|X)^2}{p_{0a}(X)^2}(D_a-p_{0a}(X))^2\\
	&\hspace{2em}+\sum_{a=0}^m\sum_{a'\neq a}\mathbb{E}[Y(a)|X]\mathbb{E}[Y(a')|X]\frac{\pi(a|X)\pi(a'|X)}{p_{0a}(X)p_{0a'}(X)}\\
	&\hspace{5em}(D_a-p_{0a}(X))(D_{a'}-p_{0a'}(X))\bigr]\\
	=&-\mathbb{E}\bigl[\sum_{a=0}^m (1-p_{0a}(X))\mathbb{E}[Y(a)|X]^2\frac{\pi(a|X)^2}{p_{0a}(X)}
	\\
	&-\sum_{a=0}^m\sum_{a'\neq a}\mathbb{E}[Y(a)|X]\mathbb{E}[Y(a')|X]\pi(a|X)\pi(a'|X)\bigr],
\end{align*}
where the last equality uses the facts that $\mathbb{E}[(D_a-p_{0a}(X))^2|X]=p_{0a}(X)(1-p_{0a}(X))$ and that $\mathbb{E}[(D_a-p_{0a}(X))(D_{a'}-p_{0a'}(X))|X]=-p_{0a}(X)p_{0a'}(X)$.
Therefore,
\begin{align*}
	&\mathbb{V}[g(Z,V^{\pi},\tilde p_0)]-\mathbb{V}[\tilde g(Z,V^{\pi},\tilde p_0)]\\
	=&\mathbb{E}\Bigl[\bigl(\sum_{a=0}^m\mathbb{E}[Y(a)|X]\frac{\pi(a|X)}{p_{0a}(X)}(D_a-p_{0a}(X))\bigr)^2\Bigr],
\end{align*}
which is nonnegative.

Since $\theta(X)=\sum_{a=0}^m\mathbb{E}[Y(a)|X]\pi(a|X)$, this is equal to
\begin{align*}
	&\mathbb{E}\Bigl[\bigl(\sum_{a=0}^m\mathbb{E}[Y(a)|X]\frac{\pi(a|X)}{p_{0a}(X)}D_a-\theta(X)\bigr)^2\Bigr]\\
	=&\mathbb{E}\Bigl[\mathbb{E}\bigl[(\sum_{a=0}^m\mathbb{E}[Y(a)|X]\frac{\pi(a|X)}{p_{0a}(X)}D_a-\theta(X))^2\bigr|X]\Bigr]\\
	=&\mathbb{E}\Bigl[\sum_{a=0}^m p_{0a}(X)\bigl(\mathbb{E}[Y(a)|X]\frac{\pi(a|X)}{p_{0a}(X)}-\theta(X)\bigr)^2\Bigr],
\end{align*}
where we use the definition of expectation for the last equality.
Under Assumption \ref{nondegeneracy}, this is greater than zero if $\Pr(\mathbb{E}[Y(a)|X]\frac{\pi(a|X)}{p_{0a}(X)}\neq \theta(X)~\text{for some}~ a)>0$.

We next compare $\ddot V^\pi$ to $\tilde V^\pi$.
Recall that $p_t=(p_{t0},...,p_{tm})'$ is the probability vector indicating the probability that each action is chosen in round $t$ and that $p=(p_0,...,p_m)'$ is a probability vector that has the same distribution as $p_t$.
We have that
\begin{align*}
	&\sqrt{T}(\ddot V^{\pi}-V^{\pi})\\
	=&\frac{1}{\sqrt{T}}\sum_{t=1}^T\bigl(\sum_{a=0}^mY_t D_{ta}\pi(a|X_t)(\frac{1}{p_{ta}}-\frac{1}{p_{0a}(X_t)})\\
	&\hspace{4em}+\sum_{a=0}^mY_t D_{ta}\frac{\pi(a|X_t)}{p_{0a}(X_t)}-V^\pi\bigr)\\
	=&\frac{1}{\sqrt{T}}\sum_{t=1}^T \bigl(\sum_{a=0}^mY_t D_{ta}\pi(a|X_t)(\frac{1}{p_{ta}}-\frac{1}{p_{0a}(X_t)})\\
	&\hspace{4em}+g(Z_t,V^{\pi},\tilde p_0)\bigr).
\end{align*}
It follows that
\begin{align*}
	&\mathbb{E}[\sum_{a=0}^mY D_a\pi(a|X)(\frac{1}{p_a}-\frac{1}{p_{0a}(X)})]\\
	=&\mathbb{E}[\mathbb{E}[\sum_{a=0}^mY D_a\pi(a|X)(\frac{1}{p_a}-\frac{1}{p_{0a}(X)})|X,p]]\\
	=&\mathbb{E}[\sum_{a=0}^m\mathbb{E}[Y(a)|X] \mathbb{E}[D_a|X,p]\pi(a|X)(\frac{1}{p_a}-\frac{1}{p_{0a}(X)})]\\
	=&\mathbb{E}[\sum_{a=0}^m\mathbb{E}[Y(a)|X] \pi(a|X)(1-\frac{p_a}{p_{0a}(X)})]\\
	=&\mathbb{E}[\mathbb{E}[\sum_{a=0}^m\mathbb{E}[Y(a)|X] \pi(a|X)(1-\frac{p_a}{p_{0a}(X)})|X]]\\
	=&\mathbb{E}[\sum_{a=0}^m\mathbb{E}[Y(a)|X] \pi(a|X)(1-\frac{\mathbb{E}[p_a|X]}{p_{0a}(X)})]\\
	=&0,
\end{align*}
where the second equality uses $Y(\cdot)\indep D|(X,p)$ and $Y(\cdot)\indep p|X$, the third equality uses $\mathbb{E}[D_a|X,p]=p_a$, and the last equality uses $p_{0a}(X)=\mathbb{E}[p_a|X]$.
By the central limit theorem, the asymptotic variance of $\sqrt{T}(\ddot V^{\pi}-V^{\pi})$ is
$$
\mathbb{E}\bigl[\bigl(\sum_{a=0}^mY D_a\pi(a|X)(\frac{1}{p_a}-\frac{1}{p_{0a}(X)})+g(Z,V^{\pi},\tilde p_0)\bigr)^2\bigr].
$$
The difference between the asymptotic variance of $\sqrt{T}(\ddot V^{\pi}-V^{\pi})$ and that of $\sqrt{T}(\tilde V^{\pi}-V^{\pi})$ is
\begin{align*}
	&\mathbb{E}\bigl[\bigl(\sum_{a=0}^mY D_a\pi(a|X)(\frac{1}{p_a}-\frac{1}{p_{0a}(X)})\bigr)^2\\
	&+2\bigl(\sum_{a=0}^mY D_a\pi(a|X)(\frac{1}{p_a}-\frac{1}{p_{0a}(X)})\bigr)g(Z,V^{\pi},\tilde p_0)\bigr].
\end{align*}
The second term equals
\begin{align*}
	&2\mathbb{E}\bigl[\bigl(\sum_{a=0}^mY D_a\pi(a|X)(\frac{1}{p_a}-\frac{1}{p_{0a}(X)})\bigr)g(Z,V^{\pi},\tilde p_0)\bigr]\\
	=&2\mathbb{E}\bigl[\bigl(\sum_{a=0}^mY D_a\pi(a|X)(\frac{1}{p_a}-\frac{1}{p_{0a}(X)})\bigr)\\
	&\hspace{10em}\bigl(\sum_{a=0}^mY D_a\frac{\pi(a|X)}{p_{0a}(X)}\bigr)\bigr]\\
	=&2\mathbb{E}\bigl[\sum_{a=0}^mY^2 D_a\frac{\pi(a|X)^2}{p_{0a}(X)}(\frac{1}{p_a}-\frac{1}{p_{0a}(X)})\bigr]\\
	=&2\mathbb{E}\bigl[\sum_{a=0}^m\mathbb{E}[Y(a)^2|X] \mathbb{E}[D_a|X,p]\frac{\pi(a|X)^2}{p_{0a}(X)}\\
	&\hspace{12em}(\frac{1}{p_a}-\frac{1}{p_{0a}(X)})\bigr]\\
	=&2\mathbb{E}\bigl[\sum_{a=0}^m\mathbb{E}[Y(a)^2|X]\frac{\pi(a|X)^2}{p_{0a}(X)}(1-\frac{p_a}{p_{0a}(X)})\bigr]\\
	=&2\mathbb{E}\bigl[\sum_{a=0}^m\mathbb{E}[Y(a)^2|X]\frac{\pi(a|X)^2}{p_{0a}(X)}(1-\frac{\mathbb{E}[p_a|X]}{p_{0a}(X)})\bigr]\\
	=&0,
\end{align*}
where the second equality uses $D_aD_{a'}=0$ for $a'\neq a$, the third equality uses $Y(\cdot)\indep D|(X,p)$ and $Y(\cdot)\indep p|X$, and the last equality uses $p_{0a}(X)=\mathbb{E}[p_a|X]$.
The first term is nonnegative, and equals
\begin{align}\label{eq:varddot}
	&\mathbb{E}\bigl[\sum_{a=0}^mY^2 D_a\pi(a|X)^2(\frac{1}{p_a^2}-\frac{2}{p_ap_{0a}(X)}+\frac{1}{p_{0a}(X)^2})\bigr]\nonumber\\
	=&\mathbb{E}\bigl[\sum_{a=0}^m\mathbb{E}[Y(a)^2|X] \pi(a|X)^2\nonumber\\
	&\hspace{6em}(\frac{1}{p_a}-\frac{2}{p_{0a}(X)}+\frac{p_a}{p_{0a}(X)^2})\bigr]\nonumber\\
	=&\mathbb{E}\bigl[\sum_{a=0}^m\mathbb{E}[Y(a)^2|X] \pi(a|X)^2\nonumber\\
	&\hspace{5em}\times(\mathbb{E}[\frac{1}{p_a}|X]-\frac{2}{p_{0a}(X)}+\frac{\mathbb{E}[p_a|X]}{p_{0a}(X)^2})\bigr]\nonumber\\
	=&\mathbb{E}\bigl[\sum_{a=0}^m\mathbb{E}[Y(a)^2|X] \pi(a|X)^2(\mathbb{E}[\frac{1}{p_a}|X]-\frac{1}{p_{0a}(X)})\bigr]
\end{align}
where the first line uses $D_aD_{a'}=0$ for $a'\neq a$, the first equality uses $Y(\cdot)\indep D|(X,p)$ and $Y(\cdot)\indep p|X$, and the last equality uses $p_{0a}(X)=\mathbb{E}[p_a|X]$.
By Jensen's inequality, $\mathbb{E}[\frac{1}{p_a}|X]-\frac{1}{p_{0a}(X)}\ge 0$.
Expression (\ref{eq:varddot}) is greater than zero if $\Pr(\mathbb{E}[Y(a)^2|X]\pi(a|X)^2(\mathbb{E}[\frac{1}{p_a}|X]-\frac{1}{p_{0a}(X)})\neq 0~\text{for some}~ a)>0$.

To show that $\tilde V^\pi_{SN}$ and $\ddot V^\pi_{SN}$ have the same asymptotic variance as $\tilde V^\pi$ and $\ddot V^\pi$, respectively, it suffices to show that the denominators of $\tilde V^\pi_{SN}$ and $\ddot V^\pi_{SN}$ converge in probability to one.
We have that
\begin{align*}
	\mathbb{E}[\sum_{a=0}^mD_a\frac{\pi(a|X)}{p_{0a}(X)}]&=\mathbb{E}[\sum_{a=0}^m\mathbb{E}[D_a|X]\frac{\pi(a|X)}{p_{0a}(X)}]\\
	&=\mathbb{E}[\sum_{a=0}^m\pi(a|X)]\\
	&=1,
\end{align*}
and that
\begin{align*}
	\mathbb{E}[\sum_{a=0}^mD_a\frac{\pi(a|X)}{p_a}]&=\mathbb{E}[\sum_{a=0}^m\mathbb{E}[D_a|X,p]\frac{\pi(a|X)}{p_a}]\\
	&=\mathbb{E}[\sum_{a=0}^m\pi(a|X)]\\
	&=1.
\end{align*}
The law of large numbers implies the convergence in probability of $\frac{1}{T}\sum_{t=1}^T\sum_{a=0}^m D_{ta}\frac{\pi(a|X_t)}{p_{0a}(X_t)}$ and $\frac{1}{T}\sum_{t=1}^T\sum_{a=0}^m D_{ta}\frac{\pi(a|X_t)}{p_{ta}}$ to one.
\qed

\end{document}